\PassOptionsToPackage{unicode}{hyperref}
\PassOptionsToPackage{hyphens}{url}
\PassOptionsToPackage{dvipsnames,svgnames,x11names}{xcolor}
\documentclass[
  12pt]{article}

\usepackage{amsmath}
\usepackage{amssymb}
\usepackage{tabularx}
\usepackage[ruled,vlined,linesnumbered]{algorithm2e}
\usepackage{amsthm}
\usepackage{mathtools}
\usepackage{multirow}
\usepackage{graphicx}

\usepackage{iftex}
\ifPDFTeX
  \usepackage[T1]{fontenc}
  \usepackage[utf8]{inputenc}
  \usepackage{textcomp} % provide euro and other symbols
\else % if luatex or xetex
  \usepackage{unicode-math}
  \defaultfontfeatures{Scale=MatchLowercase}
  \defaultfontfeatures[\rmfamily]{Ligatures=TeX,Scale=1}
\fi
\usepackage{lmodern}
\ifPDFTeX\else  
\fi
\IfFileExists{upquote.sty}{\usepackage{upquote}}{}
\IfFileExists{microtype.sty}{% use microtype if available
  \usepackage[]{microtype}
  \UseMicrotypeSet[protrusion]{basicmath} % disable protrusion for tt fonts
}{}
\makeatletter
\@ifundefined{KOMAClassName}{% if non-KOMA class
  \IfFileExists{parskip.sty}{%
    \usepackage{parskip}
  }{% else
    \setlength{\parindent}{0pt}
    \setlength{\parskip}{6pt plus 2pt minus 1pt}}
}{% if KOMA class
  \KOMAoptions{parskip=half}}
\makeatother
\usepackage{xcolor}
\makeatletter
\ifx\paragraph\undefined\else
  \let\oldparagraph\paragraph
  \renewcommand{\paragraph}{
    \@ifstar
      \xxxParagraphStar
      \xxxParagraphNoStar
  }
  \newcommand{\xxxParagraphStar}[1]{\oldparagraph*{#1}\mbox{}}
  \newcommand{\xxxParagraphNoStar}[1]{\oldparagraph{#1}\mbox{}}
\fi
\ifx\subparagraph\undefined\else
  \let\oldsubparagraph\subparagraph
  \renewcommand{\subparagraph}{
    \@ifstar
      \xxxSubParagraphStar
      \xxxSubParagraphNoStar
  }
  \newcommand{\xxxSubParagraphStar}[1]{\oldsubparagraph*{#1}\mbox{}}
  \newcommand{\xxxSubParagraphNoStar}[1]{\oldsubparagraph{#1}\mbox{}}
\fi
\makeatother

\providecommand{\tightlist}{%
  \setlength{\itemsep}{0pt}\setlength{\parskip}{0pt}}\usepackage{longtable,booktabs,array}
\usepackage{calc} % for calculating minipage widths
\usepackage{etoolbox}
\makeatletter
\patchcmd\longtable{\par}{\if@noskipsec\mbox{}\fi\par}{}{}
\makeatother
\IfFileExists{footnotehyper.sty}{\usepackage{footnotehyper}}{\usepackage{footnote}}
\makesavenoteenv{longtable}
\usepackage{graphicx}
\makeatletter
\def\maxwidth{\ifdim\Gin@nat@width>\linewidth\linewidth\else\Gin@nat@width\fi}
\def\maxheight{\ifdim\Gin@nat@height>\textheight\textheight\else\Gin@nat@height\fi}
\makeatother
\setkeys{Gin}{width=\maxwidth,height=\maxheight,keepaspectratio}
\makeatletter
\def\fps@figure{htbp}
\makeatother

\newtheorem{theorem}{Theorem}[section]

\newtheorem{condition}{Condition}
\newtheorem{remark}{Remark}[section]

\makeatletter
\@ifpackageloaded{caption}{}{\usepackage{caption}}
\AtBeginDocument{%
\ifdefined\contentsname
  \renewcommand*\contentsname{Table of contents}
\else
  \newcommand\contentsname{Table of contents}
\fi
\ifdefined\listfigurename
  \renewcommand*\listfigurename{List of Figures}
\else
  \newcommand\listfigurename{List of Figures}
\fi
\ifdefined\listtablename
  \renewcommand*\listtablename{List of Tables}
\else
  \newcommand\listtablename{List of Tables}
\fi
\ifdefined\figurename
  \renewcommand*\figurename{Figure}
\else
  \newcommand\figurename{Figure}
\fi
\ifdefined\tablename
  \renewcommand*\tablename{Table}
\else
  \newcommand\tablename{Table}
\fi
}
\@ifpackageloaded{float}{}{\usepackage{float}}
\floatstyle{ruled}
\@ifundefined{c@chapter}{\newfloat{codelisting}{h}{lop}}{\newfloat{codelisting}{h}{lop}[chapter]}
\floatname{codelisting}{Listing}

\makeatother
\makeatletter
\@ifpackageloaded{caption}{}{\usepackage{caption}}
\@ifpackageloaded{subcaption}{}{\usepackage{subcaption}}
\makeatother

\ifLuaTeX
  \usepackage{selnolig}  % disable illegal ligatures
\fi
\usepackage[]{natbib}
\usepackage{bookmark}

\IfFileExists{xurl.sty}{\usepackage{xurl}}{} % add URL line breaks if available
\hypersetup{
  pdftitle={Title},
  pdfauthor={Author 1; Author 2},
  pdfkeywords={3 to 6 keywords, that do not appear in the title},
  colorlinks=true,
  linkcolor={blue},
  filecolor={Maroon},
  citecolor={Blue},
  urlcolor={Blue},
  pdfcreator={LaTeX via pandoc}}

\newcommand{\anon}{1}

\begin{document}

\def\spacingset#1{\renewcommand{\baselinestretch}%
{#1}\small\normalsize} \spacingset{1}

%%%%%%%%%%%%%%%%%%%%%%%%%%%%%%%%%%%%%%%%%%%%%%%%%%%%%%%%%%%%%%%%%%%%%%%%%%%%%%

\if1\anon
{
  \title{\bf Heterogeneous Multisource Transfer Learning via Model Averaging for Positive-Unlabeled Data}
	\author{Jialei Liu$^{1}$, Jun Liao$^{2}$, Kuangnan Fang$^{1}$ \\
       \small $^{1}$School of Economics, Xiamen University, Fujian, China \\       
       \small $^{2}$School of Statistics, Renmin University of China, Beijing, China\\
}
	
\maketitle
\renewcommand{\thefootnote}{\fnsymbol{footnote}}
%   \footnotetext[1]{These authors contributed equally as co-corresponding authors.
% Correspondence to: Jun Liao (jliao1990@163.com) and Kuangnan Fang (xmufkn@xmu.edu.cn)}
} \fi

\if0\anon
{
  \bigskip
  \bigskip
  \bigskip
  \begin{center}
    {\LARGE\bf Title}
\end{center}
  \medskip
} \fi

\bigskip
\begin{abstract}
Positive-Unlabeled (PU) learning presents unique challenges due to the lack of explicitly labeled negative samples, particularly in high-stakes domains such as fraud detection and medical diagnosis. To address data scarcity and privacy constraints, we propose a novel transfer learning with model averaging framework that integrates information from heterogeneous data sources — including fully binary labeled, semi-supervised, and PU data sets — without direct data sharing.
For each source domain type, a tailored logistic regression model is conducted, and knowledge is transferred to the PU target domain through model averaging. Optimal weights for combining source models are determined via a cross-validation criterion that minimizes the Kullback–Leibler divergence. We establish theoretical guarantees for weight optimality and convergence, covering both misspecified and correctly specified target models, with further extensions to high-dimensional settings using sparsity-penalized estimators.
Extensive simulations and real-world credit risk data analyses demonstrate that our method outperforms other comparative methods in terms of predictive accuracy and robustness, especially under limited labeled data and heterogeneous environments.
\end{abstract}

\noindent%
\vfill

\newpage
\spacingset{1.8} % DON'T change the spacing!

\section{Introduction}\label{sec-intro}

In the field of machine learning, Positive and Unlabeled (PU) data presents a distinctive and challenging problem. PU data consists of two types of instances: positive instances and unlabeled instances. The latter may contain both positive and negative examples. One prominent application of PU learning is in financial fraud detection. In such scenarios, only a small fraction of fraudulent transactions are labeled as positive due to concealment and high verification costs, while the majority of fraudulent transactions remain undetected, blending with normal transactions in the unlabeled data. This specific data structure has driven several studies focused on PU learning techniques for financial fraud detection, such as those presented in \cite{9437749}, \cite{10020693}, and \cite{qiu2024PUFraud}. 
The scope of PU learning extends beyond financial applications. For instance, in medical diagnostics, confirmed cases are labeled as positive, while unconfirmed cases remain unlabeled. In cybersecurity, only detected attacks are labeled, whereas other potential threats go unobserved. A more comprehensive overview of PU learning applications in machine learning can be found in \cite{PUbasis2020}.

The PU structure is inherently different from the conventional supervised learning setting, which typically relies on both labeled positive and negative data. The absence of explicit negative examples in PU data complicates the learning process, as it becomes difficult to accurately distinguish between positive and negative instances without a clear reference point. 
Methodologically, recent years have seen notable progress in PU learning, with several distinct strategies emerging in the literature. The first class of methods aims to identify reliable negative samples from unlabeled data, thereby transforming the PU problem into a conventional binary classification task \citep{pu2003p&n}. The second category treats unlabeled instances as noisy negative samples, employing various noise-handling techniques. These include differential misclassification penalties that favor positive samples (\citealp{6975922}; \citealp{clasen2015pu}) and parameter tuning for balanced classification \citep{PUlogit2003}. The third distinct approach incorporates class priors through weighted loss optimization during training \citep{DuPlessis2015PU, Hsieh2015pu}.
Closely related to our work is the fourth category, exemplified by \cite{ward2009presence} and \cite{Song02012020}, which utilizes modified maximum likelihood estimation to maintain parameter interpretability while developing specialized Expectation-Maximization (EM) algorithms tailored to PU settings. Building upon these methodological foundations, a recent work by \cite{LiuPUjasa} introduced the double exponential tilting model to address distributional discrepancies between known and unknown positive samples through an EM algorithm framework. 
While these methods demonstrate strong performance in relatively large-sample regimes, its effectiveness diminishes significantly in small-sample scenarios due to the inherent difficulty in extracting reliable information from PU data.
This motivates the incorporation of auxiliary data sources to compensate for the limited information in small-sample PU data. 

\begin{figure}[htbp]
     \centering
     \includegraphics[width=\linewidth]{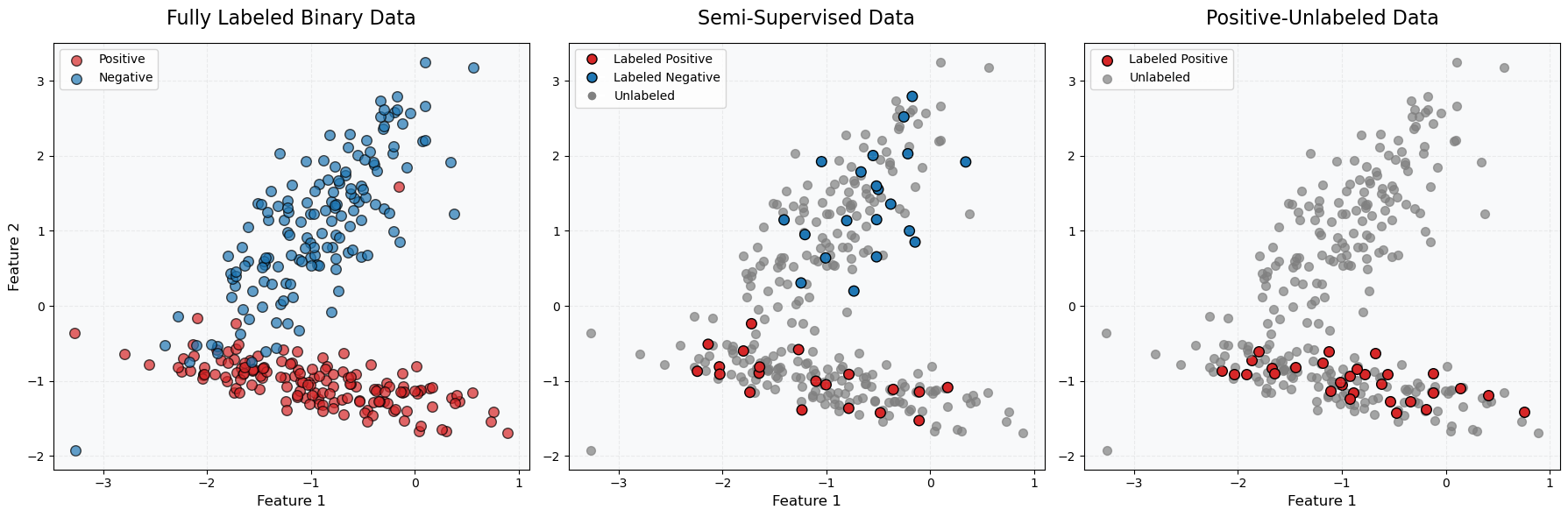}
     \caption{Comparison of the three data labeling paradigms.}
     \label{fig:data comprison}
\end{figure}

However, real-world auxiliary data often exhibits complex labeling patterns, which may include fully labeled binary-class data, semi-supervised data, and additional PU data. The distinctions among the three data types are illustrated in Figure \ref{fig:data comprison}, and their core difference lies in the degree of annotation completeness. Moreover, practical scenarios frequently encounter barriers where different data sources cannot directly share raw information due to privacy protection constraints or institutional restrictions. These cross-source information barriers create challenges for effective knowledge transfer. To address these challenges, we propose a novel cross-domain information transfer framework that incorporates auxiliary data sources to improve PU learning performance in small-sample settings.

In the context of PU learning with multiple sources under privacy constraints, the prevailing two-stage transfer learning framework \citep{Translasso2021Lisai, 10.1214/20-AOS1949, Tian02102023, Cai10102024, Zhou02012025} prove inadequate. Supervised methods are hampered by data privacy constraints that prevent source data sharing. Semi-supervised methods, whose crucial second-step parameter correction mechanism requires labeled target data, are compromised by the lack of explicit negative examples. Forcing this correction with only positive labels is likely to yield biased estimates.
Given these challenges, model averaging is an effective strategy for transfer learning with heterogeneous data sources, particularly in privacy-preserving settings. It enhances performance by aggregating model parameters to integrate diverse knowledge without data sharing, while mitigating negative transfer through an automatic down-weighting of misleading sources.
In the existing literature on model averaging for transfer learning \citep{optimal-trans,Zhang02012024}, weights are predominantly determined by minimizing prediction risk, typically quantified using squared loss. However, squared loss is unsuitable for PU data as it implicitly treats all unlabeled instances as negative examples, introducing systematic bias. As such, the Kullback-Leibler (KL) divergence serves as a commonly used and more appropriate criterion for weight selection in PU settings. While building upon the KL divergence, we introduces critical adaptations for transfer learning. Specifically, this study addresses the scenario where multiple data sources share a common feature space but exhibit distinct labeling mechanisms. Focusing on PU data as the target domain, we model the labeling processes of each source domain and estimate their corresponding parameters. These estimated parameters are then transferred to assist in label prediction for the target domain. A model-averaging-based transfer learning framework is proposed, which integrates information from heterogeneous sources via a weighted average of the transferred parameters. The weights are optimized by minimizing the KL divergence. Unlike existing KL-based weighting methods (e.g., \cite{ando&li_2017}; \cite{zou2022optimal}; \cite{yuan2024model}), we further introduce an out-of-sample KL divergence metric. This metric provides a more accurate measure of predictive capability but presents greater technical challenges. To address this, we have formulated different strategies. 

To broaden the applicability of our framework, we further extend it to high-dimensional settings via penalized estimation.
In high-dimensional settings where $p\gg n$, the existing literature on model averaging mainly adopts a two-stage approach, consisting of variable screening followed by model averaging \citep{Ando02012014,Chen03042018,He02102023}. While this approach is relatively straightforward to implement for a single data source, it imposes significant computational challenges when constructing subgroups from multiple data sources, particularly under privacy-preserving constraints. Alternatively, when $p=O(\exp(n^\alpha))$, $0<\alpha<1$, some studies apply sparsity-inducing penalties to obtain sparse parameter estimators before performing model averaging (\citealp{Zhang02042020}; \citealp{optimal-trans}). However, these studies do not provide asymptotic theory for high-dimensional settings. 
In this study, we employ an $\ell_1$ penalty to derive sparse parameter estimators, and then perform model averaging on these estimators. Additionally, we investigate the theoretical properties of the assigned weights.

In summary, we propose a novel \textbf{T}ransfer \textbf{L}earning approach based on \textbf{M}odel \textbf{A}veraging for \textbf{PU} data (TLMA-PU) that explicitly accounts for two critical considerations: (1) the heterogeneous labeling configurations across source domains, encompassing fully binary labeled, semi-supervised, and PU data types, and (2) the essential requirement for privacy-preserving data handling in cross-domain applications.
From a theoretical perspective, our work makes contributions in three aspects:

\begin{itemize}
\tightlist
    \item{\textbf{Non-Canonical Likelihood Specification}:} The likelihood function for PU observations exhibits a non-canonical form, posing challenges for model averaging methodologies. Existing theoretical analyses primarily focus on canonical likelihood formulations within the KL divergence framework. Consequently, our approach constitutes a theoretical extension. Under this setting, we establish weight optimality under model misspecification and weight consistency under correct specification for multi-source model averaging.
    \item{\textbf{Comprehensive Weight Optimality Theory}:} To our knowledge, the existing literature has not addressed out-of-sample KL divergence. Our framework fills this gap by proving weight optimality under both in-sample and out-of-sample KL divergence, offering improved applicability for transfer learning scenarios.
    \item{\textbf{High-Dimensional Theory}:} Previous theoretical analyses on model averaging have not addressed penalized estimation in high-dimensional settings. Our work extends the asymptotic theory for weighted estimators to such settings, providing new theoretical guarantees for high-dimensional PU classification problems.
\end{itemize}
The remainder of this article is organized as follows.  Section \ref{sec: TLMA-PU method} presents the proposed TLMA-PU method, detailing candidate model specification, parameter transfer process, and prediction. Section \ref{sec:Asymptotic Propertie} establishes asymptotic properties of weight estimators under both correct and incorrect model specifications. Section \ref{sec:high-dim} extends the method to high-dimensional settings, along with corresponding theoretical guarantees. Comprehensive performance evaluations are provided in Section \ref{sec: Numerical Studies}, combining simulated and real-world data. Section \ref{sec: conclusion} provides concluding remarks. All proofs are deferred to Supplement Material,  and additional numerical simulations investigating the model's performance under various data-generating scenarios are included in Supplement Material.

\section{TLMA-PU Approach}
\label{sec: TLMA-PU method}
Suppose there are $M$ source domains and a single target domain. These domains are distinguished by superscripts: source domains are indexed as 
$(m)$ for $m=1,\dots,M$, while the target domain is denoted by $(0)$. 
For the $m$-th domain, let $\boldsymbol{x}_i^{(m)}\in \mathbb R^p$ denote the covariates of subject $i$, $y_i^{(m)}\in\{0,1\}$ denote the true label, and $n_m$ denote the sample size of domain $m$. Denote $\pi_1^{(m)}=\Pr(y_i^{(m)}=1)$ as the population proportion of positive instances, and $\pi_0^{(m)}=1-\pi_1^{(m)}$ as the proportion of negative instances.

The PU data is designed as the target domain of primary interest, with the label presence defined by binary indicators $z_i^{(0)}$. For example, in the context of financial fraud detection, $z_i^{(0)}=1$ denotes labeled instances confirmed as fraudulent ($y_i^{(0)}=1$), while $z_i^{(0)}=0$ identifies unlabeled instances, containing both undetected fraudulent transactions ($y_i^{(0)}=1$) and genuine normal transactions ($y_i^{(0)}=0$).
Under this labeling scheme, the target data set comprises $n_0$ instances, each consisting of a feature vector $\boldsymbol{x}_i^{(0)}$ and binary label indicator $z_i^{(0)}$. Without loss of generality, assume that the first $n_L^{(0)}$ instances are explicitly labeled as positive ($z_i^{(0)}=1$), while the subsequent $n_U^{(0)} = n_0 - n_L^{(0)}$ instances remain unlabeled ($z_i^{(0)}=0$).

To enhance label prediction in the target domain, we leverage $M$ source data sets exhibiting heterogeneous label states. These sources are categorized into three mutually exclusive types:
\begin{itemize}
\tightlist
\item \textbf{Fully Labeled binary}: Complete data set $\{(\boldsymbol{x}_i^{(m)}, y_i^{(m)})\}_{i=1}^{n_m}$.
\item \textbf{PU}: Adopts the target domain structure: 
Labeled positive instances $\{(\boldsymbol{x}_i^{(m)}, z_i^{(m)}=1)\}_{i=1}^{n_L^{(m)}}$ and unlabeled instances $\{(\boldsymbol{x}_i^{(m)}, z_i^{(m)}=0)\}_{i=n_L^{(m)}+1}^{n_m}$ with size $n_U^{(m)} = n_m - n_L^{(m)}$.
\item \textbf{Semi-Supervised}: Comprises a labeled subset $\{(\boldsymbol{x}_i^{(m)}, y_i^{(m)}, z_i^{(m)}=1)\}_{i=1}^{n_L^{(m)}}$ with $n_{L,1}^{(m)}$ positive instances and $n_{L,0}^{(m)}$ negative instances, and an unlabeled subset $\{(\boldsymbol{x}_i^{(m)}, z_i^{(m)}=0)\}_{i=n_L^{(m)}+1}^{n_m}$ of size $n_U^{(m)} = n_m - n_L^{(m)}$.
\end{itemize}
Similar to the target domain's label indicator, the binary indicator $z_i^{(m)}$ denotes label presence for instance $i$ in the $m$-th source domain.
Building upon the defined notations, this section subsequently presents two core components:
(1) candidate model generation and parameter estimation within each domain;
(2) optimal weight estimation criterion combined with a prediction framework utilizing transferred parameters.

\subsection{Model Setup}
\label{sec:model}
For each data domain, the true labels $y_i^{(m)}$ are modeled as follows,
\begin{equation}
\text{Logit}\left(\Pr\left(y_i^{(m)}=1\big|\boldsymbol{x}_i^{(m)}\right)\right)=\boldsymbol{x}_i^{(m)\top}\boldsymbol{\beta}^{(m)},~~m=0,1,\dots,M,~~i=1,\dots,n_m.
\nonumber
\end{equation}
For domain $m$, the estimator of ${\boldsymbol{\beta}}^{(m)}$ is obtained by maximizing the likelihood function $L_n^{(m)}(\boldsymbol{\beta})$:
\[
\hat{\boldsymbol{\beta}}^{(m)} = \underset{\boldsymbol{\beta}}{\arg\max} \, L_n^{(m)}(\boldsymbol{\beta}),
\]
where $L_n^{(m)}(\boldsymbol{\beta})$ is explicitly designed according to the domain's label type.
The specific formulations are detailed below, with the target domain addressed first.

The target domain (\(m = 0\)) requires the PU-specific likelihood construction, since only presence indicators \(z_i^{(0)}\) can be observed. Following \citet{ward2009presence}, its likelihood function takes the form
\begin{align}
L_n^{(0)}(\boldsymbol{\beta}^{(0)})
&= \sum_{i=1}^{n_0}\log\left[\left( \frac{b^{(0)}\exp(\boldsymbol{x}_i^{(0)\top}\boldsymbol{\beta}^{(0)})}{1+(1+b^{(0)})\exp(\boldsymbol{x}_i^{(0)\top}\boldsymbol{\beta}^{(0)})}\right)^{z_i^{(0)}}\left(
    \frac{1+\exp(\boldsymbol{x}_i^{(0)\top}\boldsymbol{\beta}^{(0)})}{1+(1+b^{(0)})\exp(\boldsymbol{x}_i^{(0)\top}\boldsymbol{\beta}^{(0)})}\right)^{1-z_i^{(0)}} 
    \right] \nonumber \\
&= \sum_{i=1}^{n_0}\left\{z_i^{(0)}h\left(\eta_i^{(0)}\right)-g\left(\eta_i^{(0)}\right)+c\left ( z_i^{(0)},b^{(0)} \right ) \right\},
\label{likelihood:pu}
\end{align}
\vspace{-\baselineskip}
\begin{flalign*}
&\text{where } & b^{(0)} &= n_L^{(0)} / (\pi_1^{(0)} n_U^{(0)}),~~\eta_i^{(0)} = \boldsymbol{x}_i^{(0)\top} \boldsymbol{\beta}^{(0)}, &\\
&& h(t) &= t - \log(1 + \exp(t)), \\
&& g(t) &= \log(1 + (1 + b) \exp(t)) - \log(1 + \exp(t)), \\
&\text{and } & c(t, b) &= t \log b.&
\end{flalign*}

For source domains (\(m = 1,\dots,M\)), the likelihood specification is dictated by each domain's labeling mechanism, resulting in three distinct formulations.
\begin{itemize}
\tightlist
    \item \textbf{Fully labeled binary domains}: The standard logistic regression likelihood is employed:
    \begin{equation}
    L_n^{(m)}(\boldsymbol{\beta}^{(m)})=\sum_{i=1}^{n_m}\log\left[\left( \frac{\exp(\boldsymbol{x}_i^{(m)\top}\boldsymbol{\beta}^{(m)})}{1+\exp(\boldsymbol{x}_i^{(m)\top}\boldsymbol{\beta}^{(m)})}\right)^{y_i^{(m)}}\left(
    \frac{1}{1+\exp(\boldsymbol{x}_i^{(m)\top}\boldsymbol{\beta}^{(m)})}\right)^{1-y_i^{(m)}} 
    \right].
    \label{likelihood: binary}
    \end{equation}
    \item \textbf{PU domains}: The likelihood follows the same structure as the target domain: \[L_n^{(m)}(\boldsymbol{\beta})=\sum_{i=1}^{n_m}\left\{z_i^{(m)}h\left(\eta_i^{(m)}\right)-g\left(\eta_i^{(m)}\right)+c\left ( z_i^{(m)},b^{(m)} \right ) \right\},\]
    where $b^{(m)}=n_L^{(m)}/(\pi_1^{(m)} n_U^{(m)})$, $\eta_i^{(m)}=\boldsymbol{x}_i^{(m)\top}\boldsymbol{\beta}^{(m)}$, and the functions $h,g,c$ are defined as above.
    \item \textbf{Semi-supervised domains}: To leverage both labeled and unlabeled data, we utilize the likelihood formulation proposed in \cite{li2025hierarchical}:
    \begin{equation}
    \begin{aligned}
        L_n^{(m)}(\boldsymbol{\beta}^{(m)})=&\sum_{i=1}^{n_L^{(m)}}\log
        \left[\left(h_1^{(m)}\frac{\exp(\nu^{(m)}+\eta_i^{(m)})}{1+\exp(\nu^{(m)}+\eta_i^{(m)})}\right)^{y_i^{(m)}}
        \left(h_0^{(m)}\frac{1}{1+\exp(\nu^{(m)}+\eta_i^{(m)})}\right)^{1-y_i^{(m)}}
        \right]\\
        &+\sum_{i=n_L^{(m)}+1}^{n_m}\log\left[1-h_0^{(m)}\frac{1}{1+\exp(\nu^{(m)}+\eta_i^{(m)})}-h_1^{(m)}\frac{\exp(\nu^{(m)}+\eta_i^{(m)})}{1+\exp(\nu^{(m)}+\eta_i^{(m)})}\right],
    \end{aligned}
\label{likelihood:ss}
\end{equation}
%\vspace{-\baselineskip}
where $\eta_{i}^{(m)}=\boldsymbol{x}_i^{(m)\top}\boldsymbol{\beta}^{(m)}$, $h_0^{(m)}=n_{L,0}^{(m)}/(n_{L,0}^{(m)}+\pi_0^{(m)} n_U^{(m)})$, $h_1^{(m)}=n_{L,1}^{(m)}/(n_{L,1}^{(m)}+\pi_1^{(m)} n_U^{(m)})$,
and $\nu^{(m)}$ is estimated by 
\[\widehat{\nu}^{(m)}=\log\left[\left(n_{L,1}^{(m)}+\pi_{1}^{(m)}n_U^{(m)}\right)\pi_{0}^{(m)}\right]-\log\left[\left(n_{L,0}^{(m)}+\pi_{0}^{(m)}n_U^{(m)}\right)\pi_{1}^{(m)}\right].\]
\end{itemize}

\begin{remark}
Adjusting the likelihood function for different label types is necessary.
For PU learning, the standard logistic likelihood treats all unlabeled data as negative examples. This treatment shifts the decision boundary and causes severe negative bias when positive labels are scarce.
For semi-supervised learning with limited labels, models trained only on labeled data propagate labeling errors through pseudo-labels during iteration. This error propagation leads to self-reinforcing model degradation.
We address both issues using case-control correction \citep{mccullagh1989generalized}. This method leverages all available data to achieve superior parameter estimates while preserving desirable statistical properties.
\end{remark}

\begin{remark}
All three likelihood functions are derived from the logistic regression framework. Crucially, both the PU and semi-supervised paradigms rely on covariate-independent labeling mechanism. Specifically, the PU learning framework assumes the selected
completely at random (SCAR) condition where the observation probability for positive labels satisfies
\( \Pr(z_i^{(m)}=1 \mid y_i^{(m)}=1, \boldsymbol{x}_i^{(m)}) = \Pr(z_i^{(m)}=1 \mid y_i^{(m)}=1) \), which is independent of \(\boldsymbol{x}_i^{(m)}\). Similarly, semi-supervised learning assumes \( \Pr(z_i^{(m)}=1 \mid y_i^{(m)}, \boldsymbol{x}_i^{(m)}) = \Pr(z_i^{(m)}=1 \mid y_i^{(m)}) \). 
When this independence is violated, the observation probability becomes covariate-dependent and is given by \(\Pr(z_i^{(m)}=1 \mid y_i^{(m)}, \boldsymbol{x}_i^{(m)}) = f^{(m)}(\boldsymbol{x}_i^{(m)})\). This induces \(\Pr(z_i^{(m)}=1 \mid \boldsymbol{x}_i^{(m)}) = f^{(m)}(\boldsymbol{x}_i^{(m)}) \cdot \Pr(y_i^{(m)}=1 \mid \boldsymbol{x}_i^{(m)})\). In such cases, estimating $ f(\boldsymbol{x}_i^{(m)})$ is necessary to ensure the identifiability of $\boldsymbol{\beta}^{(m)}$, but this requires additional assumptions and becomes computationally complex under privacy constraints. We therefore adopt the covariate-independent labeling assumption to ensure parameter transferability across domains.
\end{remark}

\subsection{Model-Averaged Transfer Learning Procedure}
\begin{figure}[h]
    \centering
    \includegraphics[width=\linewidth]{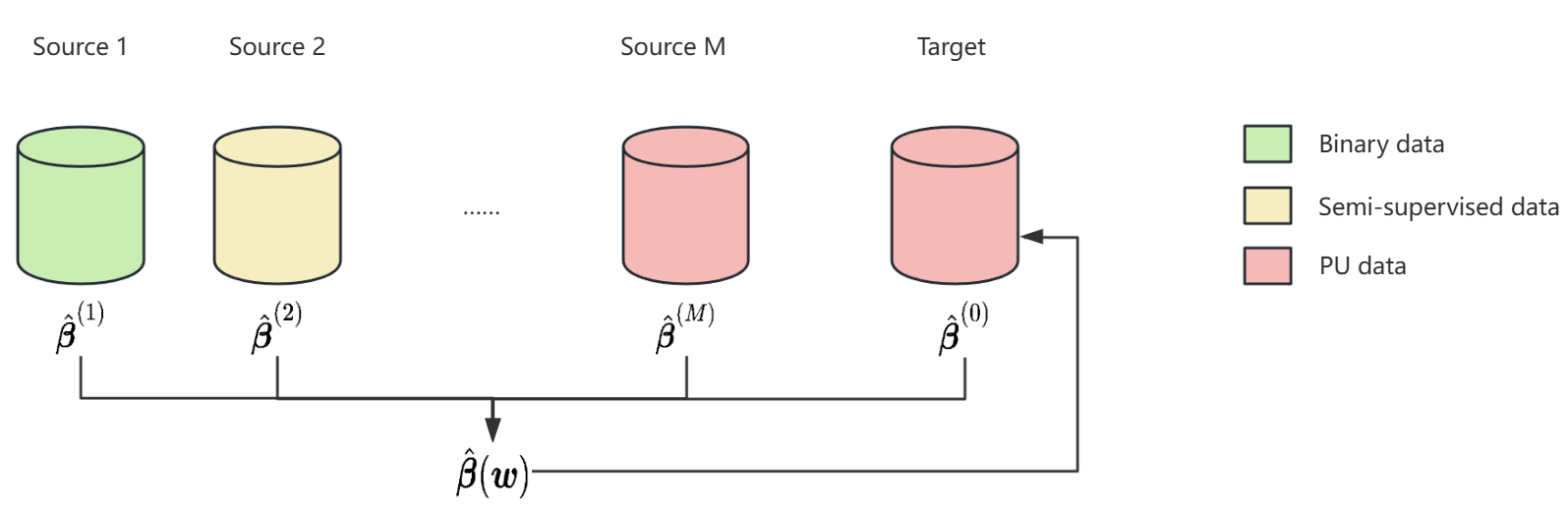}
    \vspace{-10pt}
    \caption{The process of transfer learning on PU data.}
    \label{fig:process}
\end{figure}

After obtaining \(\hat{\boldsymbol{\beta}}^{(m)}\) (\(m=0,1,\dots,M\)) via MLE, model averaging is employed to effectively transfer valuable knowledge from the source domains models to the target domain. Figure \ref{fig:process} illustrates the basic workflow of the method.
This process achieves privacy preservation by exclusively transferring parameters, without transmitting raw data or even first or second order derivatives of the objective function. By assigning appropriate weights to \(\hat{\boldsymbol{\beta}}^{(m)}\), we make predictions in the target domain. The \((M+1)\)-dimensional weight vector \(\boldsymbol{w} = (w_0, w_1, \ldots, w_M)^\top\) is constrained to the unit hypercube within \(\mathcal{R}^{M+1}\): $\mathbb{W} = \{\boldsymbol{w} \in [0,1]^{M+1} : \sum_{m=0}^{M}w_m=1\}$,
then the coefficient model averaging estimator is given by $\hat{\boldsymbol{\beta}}(\boldsymbol{w}) = \sum_{m=0}^{M} w_m \hat{\boldsymbol{\beta}}^{(m)}$. This weight allocation is essential, as it directly reflects the ``distance'' between the model of the target domain and that of the source domain, with smaller weights indicating larger ``distance''. 

To determine the optimal weights, $K$-fold $(K>1)$ cross-validation is adopted.
The labeled and unlabeled portions of the target domain data are first partitioned into 
$K$ random subsets, respectively. For each subset $\mathcal{D}_k~(k=1,\dots,K)$, the complement \(\mathcal{D}_k^c\) is used to estimate \({\boldsymbol{\beta}}^{(0)}\), referred to as \(\hat{\boldsymbol{\beta}}^{(0)}_{[\mathcal{D}_k^c]}\). Accordingly, the transfer model-averaged estimator based on the complement $\mathcal{D}_k^c$ and the sources data is then formulated as \(\hat{\boldsymbol{\beta}}^{(0)}_{[\mathcal{D}_k^c]}(\boldsymbol{w}) = w_0 \hat{\boldsymbol{\beta}}^{(0)}_{[\mathcal{D}_k^c]} + \sum_{m=1}^{M} w_m \hat{\boldsymbol{\beta}}^{(m)}.\)
Define the cross-validation criterion function 
\begin{equation} 
CV(\boldsymbol{w})=\frac{1}{n_0}\sum_{k=1}^{K}\sum_{i \in \mathcal{D}_k} \left\{-z_i^{(0)}h\left(\boldsymbol{x}_i^{(0)\top}\hat{\boldsymbol{\beta}}_{[\mathcal{D}_k^c]}(\boldsymbol{w})\right)+g\left(\boldsymbol{x}_i^{(0)\top}\hat{\boldsymbol{\beta}}_{[\mathcal{D}_k^c]}(\boldsymbol{w})\right)-c(z_i^{(0)},b^{(0)})\right\}.
\label{eq:cv}
\end{equation}
The weight estimator $\hat{\boldsymbol{w}}=(\hat{w}_0,\hat{w}_1,\dots,\hat{w}_M)$ is determined by \[\hat{\boldsymbol{w}}={\arg\min}_{\boldsymbol{w}\in \mathbb{W}} CV(\boldsymbol{w}).\] In Supplement Material, we show that the cross-validation criterion $CV(\boldsymbol{w})$ is an asymptotically unbiased estimator of the expected out-of-sample negative log-likelihood, which we seek to minimize.

For prediction, the final parameter estimator is constructed as a convex combination of the source and target domain estimators: $\hat{\boldsymbol{\beta}}(\hat{\boldsymbol{w}})=\sum_{m=0}^{M}\hat{w}_m\hat{\boldsymbol{\beta}}^{(m)}$, 
where $\hat{\boldsymbol{\beta}}^{(0)}$
is the target domain parameter vector estimated via MLE using all target samples, and $\hat{\boldsymbol{\beta}}^{(m)}$ are the estimated parameter vectors from the $M$ source domains.
Given a new observation $\boldsymbol{x}^{(0)}_{n_0+1}$ from the target domain, the predicted class probability is given by
\begin{equation}
    \Pr(\hat{y}_{n_0+1}^{(0)}=1|\boldsymbol{x}_{n_0+1}^{(0)})=\exp(\boldsymbol{x}_{n_0+1}^{^{(0)}\top}\hat{\boldsymbol{\beta}}(\hat{\boldsymbol{w}}))/\left\{ 1+\exp(\boldsymbol{x}_{n_0+1}^{^{(0)}\top}\hat{\boldsymbol{\beta}}(\hat{\boldsymbol{w}}))\right\}.
\label{eq: predict}
\end{equation}
Algorithm in Supplement Material provides a systematic outline of the aforementioned procedure.

\begin{remark}
This approach offers three principal advantages.
First, it imposes no requirement for similarity or identity in covariate distributions across domains, nor in the conditional distributions \( \Pr(y_i^{(m)}|\boldsymbol{x}_i^{(m)}) \). This aligns with practical scenarios where covariate distributions naturally exhibit heterogeneity across data sources, as do the relationships between $\boldsymbol{x}_i^{(m)}$ and $y_i^{(m)}$.
Second, the implementation of our  proposed method relies solely on parameter transfer without exchanging raw data, ensuring privacy protection, computational efficiency, and simplicity. 
Finally, it accounts for potential misspecification in all candidate models. Under some common conditions, we establish the optimality of weight estimation in Section \ref{sec:Asymptotic Propertie} in the sense of  minimizing the KL divergence.
\end{remark}

\section{Asymptotic Properties}
\label{sec:Asymptotic Propertie}
This section investigates the statistical properties of TLMA-PU under two distinct scenarios. The first scenario examines the case of potential omitted variables and target model misspecification. The second scenario considers correctly specified target models with complete variable inclusion. Together, these analyses demonstrate that the method maintains robustness against model misspecification while preserving estimation accuracy under correct specification.
In the following sections, $\|\cdot\|_0$, $\|\cdot\|_1$, $\|\cdot\|$ and $\|\cdot\|_\infty$ denote the $\ell_0$-norm, $\ell_1$-norm, $\ell_2$-norm and infinity norm, respectively. Let \(C \in \mathbb{R}^+\) denote a generic constant, which may take different values at each occurrence.

\subsection{Weight Optimality under Target Model Misspecification}
Suppose the pseudo-true value of parameter $\boldsymbol{\beta}^{(m)}$ exists, denoted as $\boldsymbol{\beta}^{*(m)}$. 
For $m=0,\dots,M$, $i=1,\dots,n_0$, define $\eta_{i,m}^{*(0)}=\boldsymbol{x}_i^{(0)\top}\boldsymbol{\beta}^{*(m)}$ and $\hat{\eta}_{i,m}^{(0)}=\boldsymbol{x}_i^{(0)\top}\hat{\boldsymbol{\beta}}^{(m)}$. Then the weighted estimators are given by 
$\eta_i^{*(0)}(\boldsymbol{w})={\textstyle \sum_{m=0}^{M}w_m \eta_{i,m}^{*(0)}}$ and $\hat{\eta}_i^{(0)}(\boldsymbol{w})={\textstyle \sum_{m=0}^{M}w_m \hat{\eta}_{i,m}^{(0)}}$.  
The binary indicator $z_i^{(0)}$ follows a Bernoulli distribution conditioned on $\boldsymbol{x}_i^{(0)}$, with probability $p_i^{(0)} = \Pr(z_i^{(0)}=1 \mid \boldsymbol{x}_i^{(0)})$, where $p_i^{(0)}$ is a function of $\boldsymbol{x}_i^{(0)}$ confined to the interval $[0, 1]$. The in-sample KL divergence between this true distribution and the logit distribution parameterized by the averaged coefficient estimator is defined as
\begin{equation}
\begin{aligned}
\text{KL}(\boldsymbol{w})&=E_z\left\{\frac{1}{n_0}\sum_{i=1}^{n_0}\left[z_{i}^{(0)}\log\frac{ p_i^{(0)}}{1-p_i^{(0)}}+\log\left(1-p_i^{(0)}\right)\right]\right.\\
&\qquad \quad \left.-\frac{1}{n_0}\sum_{i=1}^{n_0}\left[z_{ i}^{(0)}h\left(\hat{\eta}_{ i}^{(0)}(\boldsymbol{w})\right)-g\left(\hat{\eta}_{ i}^{(0)}(\boldsymbol{w})\right)+c\left ( z_{ i}^{(0)},b^{(0)} \right )\right]\right\}\\
&=\frac{1}{n_0}\sum_{i=1}^{n_0}-p_i^{(0)}h\left(\hat{\eta}_{ i}^{(0)}(\boldsymbol{w})\right)+g\left(\hat{\eta}_{ i}^{(0)}(\boldsymbol{w})\right)+a_n.
\end{aligned} 
\nonumber
\end{equation}
Here, \( a_n \) is a constant term that is independent of the weight vector \( \boldsymbol{w} \). It is important to note that in the definition of the KL divergence used here, the expectation is a conditional expectation over $z_i^{(0)}$ given $\boldsymbol{x}_i^{(0)}$. Moreover, we define $\text{KL}^*(\boldsymbol{w})
=\frac{1}{n_0}\sum_{i=1}^{n_0}[-p_{ i}^{(0)}h(\eta_{ i}^{*(0)}(\boldsymbol{w}))+g(\eta_{ i}^{*(0)}(\boldsymbol{w}))]+a_n$, retaining the constant $a_n$.
To establish the asymptotic optimality of the proposed method, we introduce the following regularity conditions. 

\begin{condition}
\label{c:unique}
There exists a unique pseudo-true value $\boldsymbol{\beta}^{*(m)}$ such that 
\begin{itemize}
\tightlist
    \item[(i)] $\boldsymbol{\beta}^{*(m)}={\arg\max}_{\boldsymbol{\beta}^{(m)}}E\left[L_n(\boldsymbol{x}^{(m)},y^{(m)};\boldsymbol{\beta}^{(m)})\right]$;
    \item[(ii)]
    $\hat{\boldsymbol{\beta}}^{(m)}-\boldsymbol{\beta}^{*(m)}=O_p\left(p^{1/2}n_m^{-1/2}\right)$,
for $m=0,1,\dots,M.$
\end{itemize}
\end{condition}

\begin{condition}
\label{c:x bounded}
$\max_{1\le i\le n_0}\max_{1\le j\le p}\left|x_{ij}\right|\le C$.
\end{condition}

\begin{condition}
\label{c: eta bound}
$\max_{m\in \{0,1,\dots,M\}}E\left[h\left({\eta}_{i,m}^{*(0)}\right)\right]^2<C$ and $\max_{m\in \{0,1,\dots,M\}}E\left[g\left({\eta}_{i,m}^{*(0)}\right)\right]^2<C$.
\end{condition}

% \begin{condition}
% \label{c: z-mu bound}
% For $i=1,\dots,n_0$, the second order moments $E\left(z_i^{(0)}-p_i^{(0)}\right)^2$ exists.
% \end{condition}

\begin{condition}
\label{c: xi and n}
 $\xi_n^{-1}pM^{1/2}\underline{n}^{-1/2}=o(1)$, where $\underline{n}=\underset{m=0,\dots,M}{\min} n_m$, and $\xi_n=\inf_{\boldsymbol{w}\in\mathbb{W}} \text{KL}^*(\boldsymbol{w})$.
\end{condition}

Condition \ref{c:unique}(i) ensures the global identifiability of $\boldsymbol{\beta}^{*(0)}$, an assumption also adopted by \cite{Song02012020} and \cite{ando&li_2017}. Condition \ref{c:unique}(ii) provides the convergence properties of the candidate model parameters, which have been utilized by \cite{Chen10122018} and \cite{ZHANG2023280} in establishing the asymptotic optimality of the weights.
Condition \ref{c:x bounded} guarantees the existence of various moments. This condition is equivalent to Equation (12) in Condition 3 of \cite{zou2022optimal}, restricting the covariates to satisfy $\max_{1\le i\le n_0}\|\boldsymbol{x}_i^{(0)}\|/p\le C$. Similar conditions can also be found in \cite{10.1214/10-AOS846} and \cite{10.1214/12-EJS731}. 
Condition \ref{c: eta bound} is a second-moment condition on both $h(\eta_{i,m}^{*(0)})$ and $g(\eta_{i,m}^{*(0)})$. 
Given the boundedness of $\eta_{i,m}^{*(0)}$ and the continuity of $h(\cdot)$, it follows that $h(\eta_{i,m}^{*(0)})$ is bounded for all $m \in {0,1,\dots,M}$. This implies that ${E}[ h(\eta_{i,m}^{*(0)})^2] < \infty$ for any $m$.
For $g(\eta_{i,m}^{*(0)})$, since $g(\cdot)$ is monotonically increasing and bounded, satisfying $0 < g(t) < b$ for all real $t$, it follows that $E[g(\eta_{i,m}^{*(0)})^2]$ is bounded for any $m$.
Condition \ref{c: xi and n} restricts the growth rate of $\inf_{\boldsymbol{w}\in\mathbb{W}} \text{KL}^*(\boldsymbol{w})$ to be not slower than $pM^{1/2}\underline{n}^{-1/2}$, implying a significant discrepancy between the target model and the true model. Similar assumptions can be found in Condition 5 of \cite{zou2022optimal} and Condition 8 of \cite{optimal-trans}.

\begin{theorem}
\label{theorem:asymptotic optimal}
Suppose Conditions \ref{c:unique}-\ref{c: xi and n} hold, then we have $\hat{\boldsymbol{w}}$ is asymptotically optimal in the sense that
\begin{equation}
    \frac{\text{KL}(\hat{\boldsymbol{w}} )}{\inf_{\boldsymbol{w}\in \mathbb{W}}\text{KL}(\boldsymbol{w})} \overset{p}{\to}1.
    \nonumber
\end{equation}
\end{theorem}
Theorem \ref{theorem:asymptotic optimal} demonstrates that the weighted estimator  is asymptotically optimal in the sense
that the KL divergence of the averaged model is asymptotically identical to that of the infeasible best model averaging estimator.
Unlike most previous model averaging approaches that rely on a single data set, our method employs candidate models based on multiple data sets. Importantly, the required conditions only necessitate a significant difference between the target model and the true model. Regardless of whether the source model is correctly specified, the weighted estimator remains asymptotically optimal.

\begin{remark}
The proof of weight optimality for the conventional exponential family \citep{ando&li_2017} does not apply to the target model, primarily due to structural differences in the link function. Recall that the canonical form of the single-sample log-likelihood is universally given by $L(\boldsymbol{\beta};y_i,\boldsymbol{x}_i)=y_i\boldsymbol{x}_i^\top\boldsymbol{\beta}-\phi(\boldsymbol{x}_i^\top\boldsymbol{\beta})+\psi(y_i)$, where $\phi(\cdot)$ is the cumulant function and $\psi(\cdot)$ is the normalizing function. Notably, the second derivative of $y_i\boldsymbol{x}_i^\top\boldsymbol{\beta}$ with respect to $\boldsymbol{\beta}$ is zero, and the conditional expectation satisfies $E(y_i|\boldsymbol{x}_i) = \phi'(\boldsymbol{x}_i^\top\boldsymbol{\beta})$.
In contrast, the single-observation PU log-likelihood, $L_{PU}(\boldsymbol{\beta};z_i,\boldsymbol{x}_i)=z_ih\left(\boldsymbol{x}_i^\top\boldsymbol{\beta}\right)-g\left(\boldsymbol{x}_i^\top\boldsymbol{\beta}\right)+c(z_i,b)$, employs a non-canonical link function. This formulation yields non-vanishing higher-order derivatives for the term $z_ih\left(\boldsymbol{x}_i^\top\boldsymbol{\beta}\right)$ with respect to $\boldsymbol{\beta}$, and satisfies 
$E(z_i|\boldsymbol{x}_i) = g'(\boldsymbol{x}_i^\top\boldsymbol{\beta})/h'(\boldsymbol{x}_i^\top\boldsymbol{\beta})$. Furthermore, the non-convexity of the PU likelihood introduces significant theoretical obstacles. Thus, to derive the asymptotic properties of the weight  estimator, we  adopt a distinct strategy to bound the difference between the KL risk and the weight choice criterion.  
\end{remark}

We further investigate the out-of-sample properties of KL divergence. For newly entered samples, let $\eta_{n_0+1,m}^{*(0)}=\boldsymbol{x}_{n_0+1}^{(0)\top}\boldsymbol{\beta}^{*(m)}$ and $\hat\eta_{n_0+1,m}^{(0)}=\boldsymbol{x}_{n_0+1}^{(0)\top}\hat{\boldsymbol{\beta}}^{(m)}$ for $m=0,\dots,M$.
The corresponding model averaging estimator $\eta_{n_0+1}^{*(0)}(\boldsymbol{w})={\textstyle \sum_{m=0}^{M}w_m \eta_{n_0+1,m}^{*(0)}}$ and $\hat\eta_{n_0+1}^{(0)}(\boldsymbol{w})={\textstyle \sum_{m=0}^{M}w_m \hat\eta_{n_0+1,m}^{(0)}}$.  
Define the out-of-sample KL divergence as
\begin{equation}
\begin{aligned}
\text{OKL}(\boldsymbol{w})&=E\left\{z_{n_0+1}^{(0)}\log\frac{p_{n_0+1}^{(0)}}{1-p_{n_0+1}^{(0)}}+\log\left(1-p_{n_0+1}^{(0)}\right)\right.\\
&\left.\qquad\quad-\left[z_{n_0+1}^{(0)}h\left(\hat{\eta}_{n_0+1}^{(0)}(\boldsymbol{w})\right)-g\left(\hat{\eta}_{n_0+1}^{(0)}(\boldsymbol{w})\right)+c(z_{n_0+1}^{(0)},b^{(0)})\right]\right\}\\
&=E\left\{-z_{n_0+1}^{(0)}h\left(\hat{\eta}_{n_0+1}^{(0)}(\boldsymbol{w})\right)+g\left(\hat{\eta}_{n_0+1}^{(0)}(\boldsymbol{w})\right)\right\}+a_n,
\end{aligned} 
\nonumber
\end{equation}
where \( a_n \) is a constant term that is independent of the weight vector \( \boldsymbol{w} \). It is important to note that the definition of OKL requires taking the expectation with respect to \( (z_{n_0+1}^{(0)}, \boldsymbol{x}_{n_0+1}^{(0)}) \). Similarly, we introduce $\text{OKL}^*(\boldsymbol{w})=E\{-z_{n_0+1}^{(0)}h({\eta}_{n_0+1}^{*(0)}(\boldsymbol{w}))+g({\eta}_{n_0+1}^{*(0)}(\boldsymbol{w}))\}+a_n$, retaining the same constant term $a_n$. To establish the asymptotic optimality of the weighted estimator under the out-of-sample KL divergence, the following conditions are introduced.
\begin{condition}
\label{c: epsilon and n}
$\epsilon_n^{-1}pM^{1/2}\underline{n}^{-1/2}=o(1)$, where $\epsilon_n=\inf_{\boldsymbol{w}\in\mathbb{W}} \text{OKL}^*(\boldsymbol{w})$.
\end{condition}
Condition \ref{c: epsilon and n} is similar to Condition \ref{c: xi and n}, restricting the growth rate of \(\inf_{\boldsymbol{w}\in\mathbb{W}} \text{OKL}^*(\boldsymbol{w})\) to be not slower than \(pM^{1/2}\underline{n}^{-1/2}\). This also indicates a significant difference between the target model and the true model.
\begin{theorem}
\label{theorem:asymptotic optimal(out-of-sample)}
Suppose Conditions \ref{c:unique}-\ref{c: eta bound} and \ref{c: epsilon and n} hold, then we have $\hat{\boldsymbol{w}}$ is asymptotically optimal in the sense that
\begin{equation}
    \frac{\text{OKL}(\hat{\boldsymbol{w}} )}{\inf_{\boldsymbol{w}\in \mathbb{W}}\text{OKL}(\boldsymbol{w})} \overset{p}{\to}1.
    \nonumber
\end{equation}
\end{theorem}
In Theorem \ref{theorem:asymptotic optimal(out-of-sample)}, we extend the asymptotic optimality to the out-of-sample context, that is, the proposed method attains asymptotic optimality in achieving the lowest possible out-of-sample KL divergence, making it more applicable to transfer learning tasks.

\subsection{Weight Convergence under Correct Target Model Specification}
This subsection considers the case where the target model is correctly specified and includes all necessary variables. A model is termed informative when its pseudo-true value satisfies $\|\boldsymbol{\beta}^{*(m)}-\boldsymbol{\beta}^{*(0)}\|=O(p^{1/2}n_m^{-1/2})$; otherwise, it is deemed uninformative.
Without loss of generality, we assume the first $M_0$ source models are uninformative, while the remaining $M-M_0$ models are informative. Let $\check{\boldsymbol{w}}=\left(\hat{w}_1,\hat{w}_2,\dots,\hat{w}_{M_0}\right)^\top$ denote the weight vector assigned to the uninformative models. Consequently, the model averaging procedure is expected to naturally discard uninformative models as the target sample size increases, resulting in $\check{\boldsymbol{w}} \overset{p}{\to} 0$. To establish this theoretical property, we impose the following additional conditions.
\begin{condition}
\label{c:b bounded}
For the target domain, there exists a positive constant $K$, such that $| \log\left(b^{(0)}\right)  | \le K$.
\end{condition}

\begin{condition}
\label{c:beta bounded}
For any $r > 0$, there exists a constant $K_r$ such that
$\max_i |\boldsymbol{x}_i^{(0)\top}\boldsymbol{\beta}| \le K_r$ a.s. for all $\boldsymbol{\beta}$ in the set $\left\{\boldsymbol{\beta}: \left\|\boldsymbol{\beta}- \boldsymbol{\beta}^{*(0)}\right\| \le r\right\}$.
\end{condition}

\begin{condition}
\label{c: hessian bounded}
Denote $\Gamma_{n}\left(\tilde{\boldsymbol{\eta}}\right)=\frac{1}{n_0}\sum_{i=1}^{n_0}\left\{-z_i^{(0)}h^{\prime\prime}\left(\tilde\eta_i^{(0)}\right)+g^{\prime\prime}\left(\tilde\eta_i^{(0)}\right)\right\}\boldsymbol{x_i}^{(0)}\boldsymbol{x_i}^{(0)\top}$, where $\tilde{\eta}_i^{(0)}$ lies between $\hat{{\eta}}_{i}^{(0)}$ and $\eta_i^{*(0)}$, $0<\sigma_{\min}\left(\Gamma_{n}\left(\tilde{\boldsymbol{\eta}}\right)\right)\le \sigma_{\max}\left(\Gamma_{n}\left(\tilde{\boldsymbol{\eta}}\right)\right)\le C <\infty$. Here, $\sigma_{\min}\left(\cdot\right)$ and $\sigma_{\max}\left(\cdot\right)$ denote the smallest and largest singular values of a matrix, respectively. 
\end{condition}

\begin{condition}
\label{c: beta_m-beta_0}
For $m=1,\dots,M_0$, the pseudo-true value ${\boldsymbol{\beta}}^{*(m)}$  of uninformative model satisfies $\left\|{\boldsymbol{\beta}}^{*(m)}-{\boldsymbol{\beta}}^{*(0)}\right\|\le C$. 
\end{condition}

\begin{condition}
\label{c: lambda(beta_m-lambda_0)}
Suppose that 
$\hat{\boldsymbol{\Delta}}=\left(\hat{\boldsymbol{\beta}}^{(1)}-\hat{\boldsymbol{\beta}}^{(0)},\dots,\hat{\boldsymbol{\beta}}^{(M_0)}-\hat{\boldsymbol{\beta}}^{(0)}\right)$ has full column rank, and $0<\sigma_{\min}\left(\hat{\boldsymbol{\Delta}}\right)\le\sigma_{\max}\left(\hat{\boldsymbol{\Delta}}\right)\le C<\infty$. 
\end{condition}

Condition \ref{c:b bounded} ensures that the proportion of positive samples does not exceed a certain threshold. A similar assumption is also mentioned in Assumption 3 of \cite{Song02012020}. Condition \ref{c:beta bounded} states that \(|\boldsymbol{x}_i^{(0)\top} \boldsymbol{\beta}|\) is bounded within a neighborhood of \(\boldsymbol{\beta}^{*(0)}\), ensuring that \((1+\exp(\boldsymbol{x}_i^{(0)\top} \boldsymbol{\beta}))^{-1}\) remains between 0 and 1. The similar assumption is also found in Assumption 2 of \cite{Song02012020}. 
Condition \ref{c: hessian bounded} restricts the singular values of the second derivative of the objective function to be bounded. To illustrate the reasonability of Condition \ref{c: hessian bounded}, consider two simple cases: (1) when the data set contains only unlabeled samples, then $n_L^{(0)}=0$, $b^{(0)}=0$ and $z_i^{(0)}=0$ for $i=1,\dots,n_0$, causing $\Gamma_{n}\left(\tilde{\boldsymbol{\eta}}\right)$ to reduce to a zero matrix with all singular values being zero; 
(2) similarly, when the data set consists entirely of positive samples, i.e., $b^{(0)}\to \infty$ and $z_i^{(0)}=1$ for $i=1,\dots,n_0$, $\Gamma_{n}\left(\tilde{\boldsymbol{\eta}}\right)$ also becomes a zero matrix with all singular values equal to zero. Between these two extremes, the PU data scenario ensures that under Conditions \ref{c:b bounded} and \ref{c:beta bounded}, $\Gamma_{n}\left(\tilde{\boldsymbol{\eta}}\right)$ remains a nonsingular symmetric matrix, thereby guaranteeing bounded singular values. 
Condition \ref{c: beta_m-beta_0} restricts the distance between the uninformative model parameter estimator $\boldsymbol{\beta}^{(m)}$ and $\boldsymbol{\beta}^{(0)}$ to be finite. Combined with Condition \ref{c:x bounded}, this yields results analogous to Condition 10 in \cite{optimal-trans}. Combining Conditions \ref{c:unique} and \ref{c: beta_m-beta_0}, $\|\hat{\boldsymbol{\beta}}^{(m)}-\hat{\boldsymbol{\beta}}^{(0)}\|\le\|\hat{\boldsymbol{\beta}}^{(m)}-{\boldsymbol{\beta}}^{*(m)}\|+\|\hat{\boldsymbol{\beta}}^{(0)}-{\boldsymbol{\beta}}^{*(0)}\|+\|{\boldsymbol{\beta}}^{*(m)}-{\boldsymbol{\beta}}^{*(0)}\|=O_p(1)$. 
Condition \ref{c: lambda(beta_m-lambda_0)} restricts the relationship between $\hat{\boldsymbol{\beta}}^{(m)}$ and $\hat{\boldsymbol{\beta}}^{(0)}$ across different uninformative sources, which is similar to C9  of \cite{yuan2024model}. Specifically, when the matrix $\hat{\boldsymbol{\Delta}}$ is of full column rank, $\hat{\boldsymbol{\Delta}}^\top\hat{\boldsymbol{\Delta}}$ becomes a symmetric positive definite matrix, ensuring that its minimum singular value remains bounded away from zero. Additionally, the maximum singular value of $\hat{\boldsymbol{\Delta}}^\top\hat{\boldsymbol{\Delta}}$ is also constrained to be finite. 

\begin{theorem}
\label{theorem:weights convergence}
Suppose the target model is correctly specified both in terms of the model form and covariate completeness, if Conditions \ref{c:unique}-\ref{c:x bounded} and \ref{c:b bounded}-\ref{c: lambda(beta_m-lambda_0)} hold, then
\begin{equation}
\left\|\check{\boldsymbol{w}}\right\|\overset{p}{\to}0.
\label{the: weight convergence}
\end{equation}
\end{theorem}
Theorem \ref{theorem:weights convergence} demonstrates that the model weights associated with uninformative models will asymptotically converge to 0, while the optimal weights allocation concentrates on the informative models. Furthermore, the method remains robust to shifts in covariate distributions across domains, as it requires no explicit distributional assumptions on the covariates.

\section{Transfer Learning by Model Averaging in High-Dimensional Settings}
\label{sec:high-dim}
In this section, a model averaging framework for transfer learning under high dimensional settings is established. In such settings, the $\ell_1$ penalty can be imposed to achieve sparse coefficient estimation and enable knowledge transfer. Specifically, for the $m$-th data set, 
\begin{equation}
\hat{\boldsymbol{\beta}}^{(m)}=\underset{\boldsymbol{\beta}}{\arg\min}\left\{-L_n^{(m)}(\boldsymbol{\beta})+\lambda^{(m)}\|\boldsymbol{\beta}\|_1\right\},
\nonumber
\end{equation}
where $\lambda^{(m)}$ is a tuning parameter and $L_n^{(m)}(\boldsymbol{\beta})$ is defined as in Section \ref{sec: TLMA-PU method}. This likelihood $L_n^{(m)}(\boldsymbol{\beta})$ admits distinct representations across data types. 
The subsequent knowledge transfer procedure remains unchanged, with the optimal weights estimated by $\hat{\boldsymbol{w}}={\arg\min}_{\boldsymbol{w}\in \mathbb{W}} CV(\boldsymbol{w})$, consistent with prior procedure, where
\begin{equation} 
CV(\boldsymbol{w})=\frac{1}{n_0}\sum_{k=1}^{K}\sum_{i \in \mathcal{D}_k} \left\{-z_i^{(0)}h\left(\boldsymbol{x}_i^{(0)\top}\hat{\boldsymbol{\beta}}_{[\mathcal{D}_k^c]}(\boldsymbol{w})\right)+g\left(\boldsymbol{x}_i^{(0)\top}\hat{\boldsymbol{\beta}}_{[\mathcal{D}_k^c]}(\boldsymbol{w})\right)-c(z_i^{(0)},b^{(0)})\right\}.
\nonumber
\end{equation}

For the asymptotic property, the aforementioned theorem can be further extended to high-dimensional settings. Let $S_m\subseteq\{1,\dots,p\}$ denote the support set of $\boldsymbol{\beta}^{*(m)}$, and $s_m=\|{\boldsymbol{\beta}}^{*(m)}\|_0=\sum_{j=1}^p\mathbb{I}(\beta_j^{*(m)}\ne0)\ll p$, which corresponds to the intrinsic sparsity of the candidate model. For the model averaging estimator, the support $S\subseteq\cup_{m=0}^{M}S_m$, and the cardinality satisfies $|S|\le(M+1)\overline{s}\ll p$, where $\overline{s}={\max}_{m} s_m$. 
Furthermore, a random vector $\boldsymbol{x}$ is said to follow a sub-Gaussian distribution with parameter $\sigma_x^2 > 0$ if for any unit vector $\boldsymbol{u} \in \mathbb{R}^p$, the real random variable $\boldsymbol{u}^\top\boldsymbol{x}$ is sub-Gaussian. Specifically, for all $t \in \mathbb{R}$, $E\{\exp\{t\boldsymbol{u}^\top[\boldsymbol{x}-E(\boldsymbol{x})]\}\}\le \exp(t^2\sigma_x^{2}/2)$.  
Additionally, the following conditions are needed. 

\begin{condition}
\label{c: hd regime}
Assume \(\sqrt{\bar{s}\log p / \underline{n} }= o(1)\).
\end{condition}

\begin{condition}
\label{c: x sub-gaussian}    
The covariates ${\boldsymbol{x}_i^{(0)}}\in\mathbb{R}^p$ are independent and identically distributed (i.i.d.) sub-Gaussian random vectors with zero mean. Moreover, the covariance matrix $\boldsymbol{\Sigma}^{(0)} = E(\boldsymbol{x}_i^{(0)}\boldsymbol{x}_i^{(0)\top})$ is positive definite, and there exist constants $0 < C_1 < C_2 < \infty$ such that $C_1 \le \sigma_{\min}(\boldsymbol{\Sigma}^{(0)})<\sigma_{\max}(\boldsymbol{\Sigma}^{(0)}) \le C_2$. 
\end{condition}

\begin{condition}
\label{c:beta high dim}
The estimator $\hat{\boldsymbol{\beta}}^{(m)}$ and the pseudo-true value ${\boldsymbol{\beta}}^{*(m)}$ satisfies 
\begin{itemize}
\tightlist
\item[(i)] 
$\|\hat{\boldsymbol{\beta}}^{(m)}-\boldsymbol{\beta}^{*(m)}\|=O_p\left(\sqrt{s_m\log p/n_m}\right)$
for $m=0,1,\dots,M$;
\item[(ii)] $\|\hat{\boldsymbol{\beta}}^{(m)} - {\boldsymbol{\beta}}^{*(m)}\|_1=O_p(s_m\sqrt{\log p/n_m})$ for $m=0,1,\dots,M$;
\item[(iii)] $\max_j \max_m\left|\beta_j^{*(m)}\right|< C<\infty$. 
\end{itemize}
\end{condition}

\begin{condition}
\label{c: xi&n&p&m(high_dim)}
$\xi_p^{-1}\underline{n}^{-1/2}M\bar{s}\log p=o(1)$, where $\xi_p=\inf_{\boldsymbol{w}\in\mathbb{W}} \text{KL}^*(\boldsymbol{w})$.
\end{condition}

\begin{condition}
\label{c: xi&n&p&m_OKL(high_dim)}
$\epsilon_p^{-1}\underline{n}^{-1/2}M\bar{s}\log p=o(1)$, where $\epsilon_p=\inf_{\boldsymbol{w}\in\mathbb{W}} \text{OKL}^*(\boldsymbol{w})$.
\end{condition}
 
Condition \ref{c: hd regime} posits a sparsity assumption. It implies that the rate $\sqrt{s_m\log p/n_m}=o(1)$. This rate is common in literature and is necessary to establish asymptotically theory for high-dimensional settings. 
Condition \ref{c: x sub-gaussian} assumes that the covariates $\boldsymbol{x}_i^{(0)}$ follow a sub-Gaussian distribution, which is a common assumption in high-dimensional statistical analysis, as employed in \cite{Song02012020} and \cite{Tian02102023}. 
Additionally, we also assume the expected covariance matrix is positive definite with bounded eigenvalues, which is also commonly used in high-dimensional literature (\citealp{10.1214/14-AOS1221}; \citealp{JMLR:v22:19-132}). 

Condition \ref{c:beta high dim}(i) specifies the convergence rate of estimators under the $\ell_1$ penalty following \cite{JMLR:v16:loh15a}. Moreover, \cite{Song02012020} verified the convergence rate of $\|\hat{\boldsymbol{\beta}}^{(m)}-\boldsymbol{\beta}^{*(m)}\|$ under the PU loss with $\ell_1$ penalty, where the error bound is linear in $\sqrt{s_m\log p/n_m}$ with varying $s_m$ and $n_m$.
Condition \ref{c:beta high dim}(ii) is a common property in high-dimensional settings. The analysis of such conditions often relies on a cone of the form $C(S,c_0)=\{\boldsymbol{v}\in\mathbb{R}^p:\|\boldsymbol{v}_{S^c}\|_1\le c_0\|\boldsymbol{v}_S\|_1\}$, where $\boldsymbol{v}_S$ is the projection of $\boldsymbol{v}$ onto the subspace indexed by $S$ and $c_0$ is a positive constant. Under the mild assumption that $\hat{\boldsymbol{\beta}}^{(m)}-\boldsymbol{\beta}^{*(m)}\in C(S_m,c_0)$, we have $\|\hat{\boldsymbol{\beta}}^{(m)}-\boldsymbol{\beta}^{*(m)})\|_1=\|(\hat{\boldsymbol{\beta}}^{(m)}-\boldsymbol{\beta}^{*(m)})_{S_m}\|_1+\|(\hat{\boldsymbol{\beta}}^{(m)}-\boldsymbol{\beta}^{*(m)})_{S_m^c}\|_1\le (c_0+1)\|(\hat{\boldsymbol{\beta}}^{(m)}-\boldsymbol{\beta}^{*(m)})_{S_m}\|_1\le(c_0+1)\sqrt{s_m}\|\hat{\boldsymbol{\beta}}^{(m)}-\boldsymbol{\beta}^{*(m)}\|=O_p(s_m\sqrt{\log p/n_m})$. This indicates that under the aforementioned assumption, Condition \ref{c:beta high dim}(ii) can be derived from Condition \ref{c:beta high dim}(i).
Condition \ref{c:beta high dim}(iii) imposes an additional requirement that each component of the pseudo-true parameter across all candidate models is bounded in absolute value. This aligns with the practice in the literature of constraining the parameter space to a bounded domain $\mathbb{B}= \{ \boldsymbol{\beta} \in \mathbb{R}^p : \|\boldsymbol{\beta}\|^2 \leq R \}$ for some $R > 0$ \citep{Song02012020}, which inherently ensures component-wise boundedness since $\max_{1 \leq j \leq p} |\beta_j^{*(m)}| \leq \|\boldsymbol{\beta}^{*(m)}\| \leq R$ for $\boldsymbol{\beta}^{*(m)}\in \mathbb{B}$.

Conditions \ref{c: xi&n&p&m(high_dim)} and \ref{c: xi&n&p&m_OKL(high_dim)} are analogous to Conditions \ref{c: xi and n} and \ref{c: epsilon and n}, respectively. These conditions impose restrictions to ensure significant differences between the target model and the true model. 
We use $\asymp$ to indicate asymptotically equivalent. Assume that $\xi_p \asymp \underline{n}^{-\alpha}$ for some $\alpha\ge0$, $p \asymp \exp(\underline{n}^{\gamma})$ for some $0<\gamma<1$ and $\overline{s}=O(1)$. Given that the number of candidate models is limited, the condition simplifies to $\underline{n}^{\alpha+\gamma-1/2} = o(1)$. This requirement implies $\alpha+\gamma<1/2$. If we assume $p\asymp \underline{n}^\gamma$ for some $\gamma\ge1$ and $\overline{s}=O(\log p)$, then the condition changes to $n^{\alpha-1/2}(\log n)^{2}=o(1)$, which requires $0\le\alpha<1/2$.
Identical analysis applies to 
$\epsilon_p$ upon substitution for $\xi_p$. 

\begin{theorem}
\label{theorem:asymptotic optimal (high dim)}
Suppose the target model is misspecified and Conditions \ref{c: eta bound}, \ref{c: hd regime}-\ref{c: xi&n&p&m(high_dim)} hold, then we have $\hat{\boldsymbol{w}}$ is asymptotically optimal in the sense that
\begin{equation}
    \frac{\text{KL}(\hat{\boldsymbol{w}} )}{\inf_{\boldsymbol{w}\in \mathbb{W}}\text{KL}(\boldsymbol{w})} \overset{p}{\to}1.
    \nonumber
\end{equation}
\end{theorem}

\begin{theorem}
\label{theorem:asymptotic optimal(out-of-sample)(high dim)}
Suppose target model is misspecified and Conditions \ref{c: eta bound}, \ref{c: hd regime}-\ref{c:beta high dim} and \ref{c: xi&n&p&m_OKL(high_dim)} hold, then we have $\hat{\boldsymbol{w}}$ is asymptotically optimal in the sense that
\begin{equation}
    \frac{\text{OKL}(\hat{\boldsymbol{w}} )}{\inf_{\boldsymbol{w}\in \mathbb{W}}\text{OKL}(\boldsymbol{w})} \overset{p}{\to}1.
    \nonumber
\end{equation}
\end{theorem}
These two theorems demonstrate the asymptotic optimality under model misspecification in high-dimensional settings. 
To establish weight convergence under the correctly specified target model in high-dimensional settings, we define a source model as informative if its pseudo-true parameter satisfies $\|\boldsymbol{\beta}^{*(m)}-\boldsymbol{\beta}^{*(0)}\|=O(\sqrt{s_m\log p/n_m})$; otherwise, it is uninformative. We subsequently require another condition to establish weight convergence under correctly specified target model.
\begin{condition}
\label{c: sparsity(weight converge)}
$\sqrt{{\bar{s}p^2\log p}/{\underline{n}^{3}}}=o(1)$.
\end{condition}
Condition \ref{c: sparsity(weight converge)} allows that the dimensionality grows as $p \asymp \underline{n}^{\gamma}$ for some $1 \leq \gamma < 3/2$, and that the sparsity level satisfies $\bar{s} = O(\log p)$. 
Under the current framework, this polynomial scaling allows for a relatively direct proof. In contrast, establishing weight convergence theory under exponential scaling $p\asymp\exp(\underline{n}^{\alpha})$ for some $\alpha>0$, would necessitate a broader set of technical assumptions.
\begin{theorem}
\label{theorem:weights convergence(high_dim)}
Suppose the target model is correctly specified both in terms of the model form and covariate completeness, if Conditions \ref{c:b bounded}-\ref{c:beta bounded}, \ref{c: beta_m-beta_0}-\ref{c: lambda(beta_m-lambda_0)}, \ref{c: x sub-gaussian}-\ref{c:beta high dim} and \ref{c: sparsity(weight converge)} hold, then
\begin{equation}
\left\|\check{\boldsymbol{w}}\right\|\overset{p}{\to}0.
\nonumber
\end{equation}
\end{theorem}
Theorem \ref{theorem:weights convergence(high_dim)} establishes that, in sparse high-dimensional settings, the weights assigned to uninformative models converge asymptotically to zero. In contrast to most previous model averaging studies, which typically focus on proving that the weights on informative models sum to one, our proof strategy directly leverages the definition of the weight estimator minimizing $CV(\boldsymbol{w})$. This approach allows us to derive weight convergence without imposing restrictions on the risk of the model averaging estimator under misspecified models. Moreover, Theorem \ref{theorem:weights convergence(high_dim)} extends the convergence results of Theorem \ref{theorem:weights convergence} to high-dimensional regimes meeting the sparsity requirement in Condition \ref{c: sparsity(weight converge)}, thereby filling a gap in the literature on weight convergence in model averaging with sparse coefficient vectors.

\section{Numerical Studies}
\label{sec: Numerical Studies}
In this section, simulated and empirical data are utilized to validate the effectiveness of the proposed method. 

\subsection{Simulation}
\label{sec:simulation}
In this subsection, numerical simulations are conducted to explore the applicability of TLMA-PU in complex situations. Consistent with \cite{ZHANG2023280}, the proposed method maintains robust finite-sample performance across different cross-validation fold choices. We employed 5-fold cross-validation ($K=5$) to determine the weights. Three different scenarios are considered, with the data generation and simulation settings outlined as follows.
\begin{itemize}
\tightlist
    \item[\textbf{Case 1}] (Omitted-variable misspecification) To evaluate the performance of the proposed method under parameter misspecification, we conducted simulations across seven data domains: binary labeled domains ($m=1,2$), PU domains ($m=3,4,0$), and semi-supervised domains ($m=5,6$), where $m=0$ represents the target domain and the others are source domains.
    It is assumed that the ten variables $\boldsymbol{x}_i^{(m)}$ come from $N(0,\boldsymbol{\Sigma})$ with $\boldsymbol{\Sigma}_{ij}=0.3^{|i-j|}$, $\Pr(y_i^{(m)}=1|\boldsymbol{x}_i^{(m)},\boldsymbol{\beta}^{\text{true}})=\exp(\boldsymbol{x}_i^{{(m)}\top}\boldsymbol{\beta}^{\text{true}})/(1+\exp(\boldsymbol{x}_i^{{(m)}\top}\boldsymbol{\beta}^{\text{true}}))$, and $y_i^{(m)}\sim \text{Bernoulli}(\Pr(y_i^{(m)}=1|\boldsymbol{x}_i^{(m)},\boldsymbol{\beta}^{\text{true}}))$, where the true coefficients are set in three scenarios,
    \begin{equation}
    \boldsymbol{\beta}_1^{\text{true}}=(\underset{5}{\underbrace{1,\dots,1}} ,\underset{p-5}{\underbrace{0,\dots,0}} ), \quad
    \boldsymbol{\beta}_2^{\text{true}}=(\underset{p}{\underbrace{{0,1},\dots,{0,1}}}),\quad
    \boldsymbol{\beta}_3^{\text{true}}=(\underset{5}{\underbrace{1,\dots,1}},\underset{p-4}{\underbrace{0,\dots,0}},0.5 ).
    \nonumber
    \end{equation} 
    Models $m=1,3,5$ utilize $\boldsymbol{\beta}_1^{\text{true}}$ as the true parameter, models $m=2,4,6$ adopt $\boldsymbol{\beta}_2^{\text{true}}$ as the true parameter, and the target model $m=0$ employs $\boldsymbol{\beta}_3^{\text{true}}$ as the true parameter. 
    By excluding the final covariate from all candidate models, we simulate a scenario where all candidate models are misspecified. 
    
    \item[\textbf{Case 2}] (Correct specification) To evaluate the model's behavior under correct specification, we conducted simulations based on the following data-generating process. Ten data sets are considered. We set $m=1,2,3$ as binary data domain, $m=4,5,6,0$ as PU data domain, and $m=7,8,9$ as semi-supervised data domain; let 
    \begin{equation}
    \boldsymbol{\beta}_4^{\text{true}}=(\underset{5}{\underbrace{1,0,1,0,1}} ,\underset{p-5}{\underbrace{0,\dots,0}} ), \quad
    \boldsymbol{\beta}_5^{\text{true}}=(\underset{p-5}{\underbrace{0,\dots,0}} ,\underset{5}{\underbrace{1,1,1,1,1}}  ),
    \nonumber
    \end{equation} 
    where the true parameter of domains 0,1,4,7 takes value of $\boldsymbol{\beta}_1^{\text{true}}$, the true parameter of domains 2,5,8 takes value of $\boldsymbol{\beta}_4^{\text{true}}$, and the true parameter of domains 3,6,9 takes value of $\boldsymbol{\beta}_5^{\text{true}}$. All the other settings are the same as in Case 1.
    
    \item[\textbf{Case 3}] (Link function misspecification) Additionally, we examine the case of misspecified but informative target models. The parameter settings are the same as in Case 2, but the data generation utilizes the probit model, i.e., $\Pr(y_i^{(m)}=1|\boldsymbol{x}_i^{(m)},\boldsymbol{\beta}^{\text{true}})=\Phi(\boldsymbol{x}_i^{{(m)}\top}\boldsymbol{\beta}^{\text{true}})$, where $\Phi(\cdot)$ is the cumulative distribution function of the standard normal distribution.
\end{itemize}

A total of 100,000 samples are randomly generated to represent the population. For data sets with complete labels, $n_m = 400, 800, 1600$ samples are drawn. For data sets with incomplete labels, the number of labeled samples is set as $n_L^{(m)} = n_m \times p_L$, where $p_L\in \{0.3, 0.4\}$. Specifically, for PU-type data, $n_L^{(m)}$ samples with label 1 are drawn, and $n_m - n_L^{(m)}$ samples are drawn from the remaining samples as unlabeled. For semi-supervised data, $n_m$ samples are drawn, with $n_L^{(m)}$ samples retaining their labels and the rest having labels masked. 
For each scenario, we conducted $R=100$ replications and compared the performance of five methods for handling PU data, where the first two methods utilize a single data set, while the latter three methods employ multiple data sets. Specifically, the following methods are considered:
\begin{itemize}
\tightlist
    \item[1)] Single-PU: This approach applies MLE via Equation (\ref{likelihood:pu}) for single target parameter estimation, without utilizing source domain information.
    \item[2)] Oracle: The method assumes that the true values of all labels are accessible and employs the complete set of $\{y_i^{(0)}\}_{i=1}^{n_0}$ to construct a true model. Specifically, for Case 1 and Case 2, the logit model is utilized, while for Case 3, the probit model is employed. Although this approach is not feasible in practical applications, its performance metrics can serve as a comparative gold standard.
    \item[3)] Translasso: The two step transfer learning method  \citep{Translasso2021Lisai,Tian02102023}. In practice, we integrate all source data into a single data set and treat unlabeled samples as negative instances for incomplete labeling cases. This violates the restrictions on data exchange between data sources.
    \item[4)] Equal-Weighted: Equal weights are assigned to each model in the model averaging.
    \item[5)] TLMA-PU: The proposed transfer learning by model averaging for PU data.
\end{itemize}

We set the test sample size $n_\text{test}=500$, maintaining the same ratio of labeled to unlabeled data as in the training set, and evaluate the model's performance on test samples using the following metrics: Accuracy (ACC), Area Under the Curve (AUC), Adjusted Area Under the Curve (AUC\_adj) \citep{Song02012020}, True Positive Rate (TPR), False Positive Rate (FPR), and relative KL loss (RKL), where ACC, AUC, TPR, FPR require truth labels for computation.
The AUC\_adj is defined as:
\begin{equation}
\text{AUC\_{adj}} = \left(\text{AUC\textsubscript{naive}} - \pi_1^{(0)}/2\right)/\left(1 - \pi_1^{(0)}\right),
\label{metric:AUC_adj}
\end{equation}
where $\text{AUC\_{naive}}$ denotes the AUC calculated using PU labels $\{z_i^{(0)}\}_{i=1}^{n_{\text{test}}}$, 
and $\pi_1^{(0)}$ represents the target class prior.
The RKL is defined as:
\begin{equation}
\text{RKL}=\left(\sup_{\boldsymbol{\beta}}L_n^{(0)}(\boldsymbol{\beta})-L_n^{(0)}(\hat{\boldsymbol{\beta}})\right)\bigg/n_\text{test},
\label{metric:RKL}
\end{equation}
where $\widehat{\boldsymbol{\beta}}$ denotes the estimated coefficients, 
$L_n^{(0)}(\boldsymbol{\beta})$ is the PU likelihood function for target test samples, 
and $\sup_{\boldsymbol{\beta}} L_n^{(0)}(\boldsymbol{\beta})$ represents the maximum achievable likelihood. 
This metric is equivalent to the difference between the KL divergence and its minimum value. For clarity, the RKL is presented after being multiplied by 100.
All the results are presented in Tables S1-S3 in Supplement Material. 
The simulation results demonstrate that our proposed method performs effectively. It significantly outperforms the Single-PU by leveraging multi-source data. Even with limited sample sizes, our method achieves higher out-of-sample prediction accuracy than the Oracle method, as the latter relies solely on a single target data set for modeling, whereas our approach incorporates valuable information from multiple auxiliary sources. 
Furthermore, when compared to Translasso, our method exhibits superior adaptability to the PU learning scenario. Finally, unlike Equal-Weighted approach, TLMA-PU employs an adaptive weighting strategy that assigns greater importance to informative sources, highlighting the effectiveness of its design.

\begin{figure}[H]
    \centering
    \includegraphics[width=0.52\linewidth]{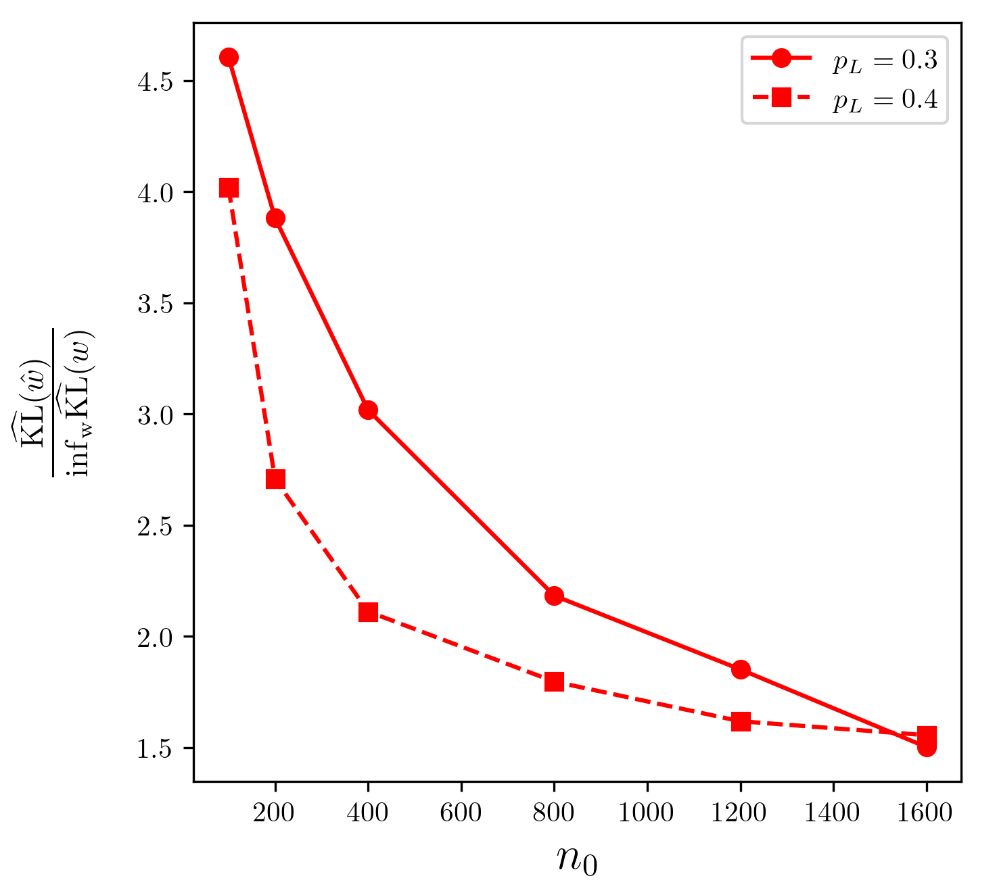}
    \caption{$\widehat{\text{KL}}(\hat{\boldsymbol{w}})/\inf_{w\in\mathbb{W}}\widehat{\text{KL}}(\boldsymbol{w})$ of Case 1 under different training sample sizes.}
    \label{fig:sim-KL}
\end{figure}

What's more, we calculate the estimated KL divergence for Case 1, i.e.,
\[\widehat{\text{KL}}(\boldsymbol{w})=\frac{1}{n^*}{\sum_{i=1}^{n^*}\left\{z_{i}^{(0)}\left[h\left(\eta_{i}^{*(0)}\right)-h\left(\hat{\eta}_{i}^{(0)}(\boldsymbol{w})\right)\right]-\left[g\left(\eta_{i}^{*(0)}\right)-g\left(\hat{\eta}_{i}^{(0)}(\boldsymbol{w})\right)\right]\right\}},\]
where $n^{*}$ is set to 3000. 
Figure \ref{fig:sim-KL} plots $\widehat{\text{KL}}(\hat{\boldsymbol{w}})/\inf_{w\in\mathbb{W}}\widehat{\text{KL}}(\boldsymbol{w})$ under different training sample sizes. The plot shows that the ratio tends to approach 1 as the target sample size increases, which is consistent with the weight optimality theorems.

\begin{figure}[H]
    \centering
    \includegraphics[width=0.5\linewidth]{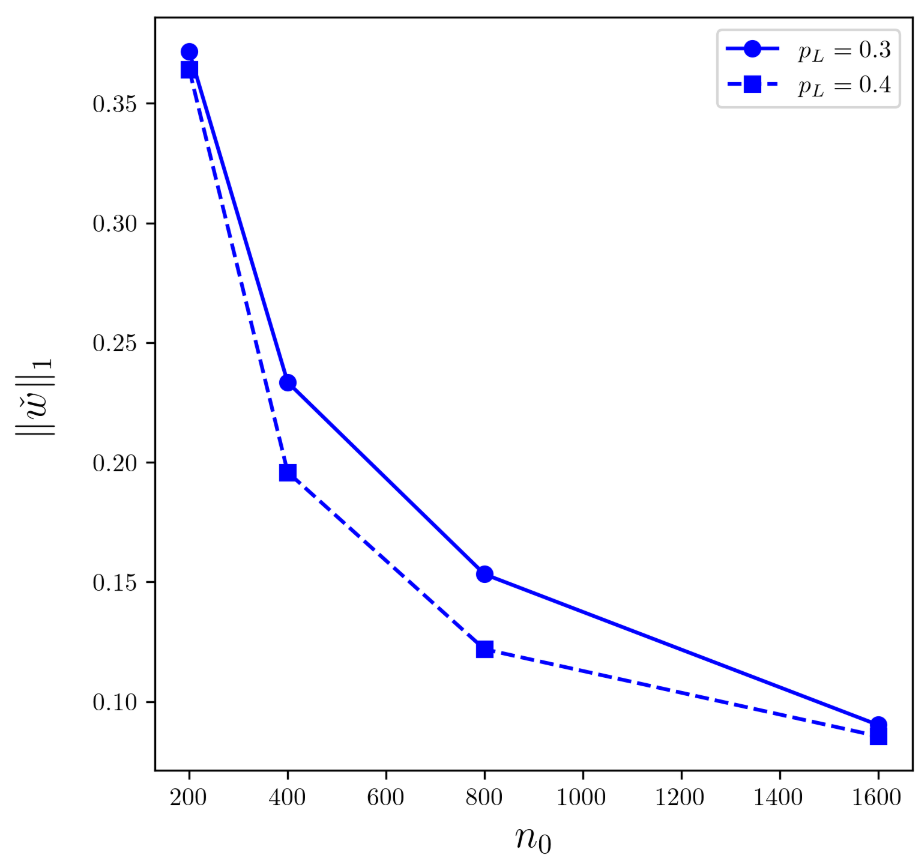}
    \caption{The sum of weights assigned to uninformative models in Case 2.}
    \label{fig:sum of weights}
\end{figure}

We also illustrate the weight allocation patterns for Cases 2 and 3, which can be founded in Supplement Material.
In Case 2, the candidate model set contains a correctly specified model that captures the true data-generating process. The resulting weights concentrate on informative models, with weights for non-informative models asymptotically approaching zero as the sample size increases.
In Case 3, the true models are probit whereas the working models are logit. This mismatch leads to misspecification of all candidate models. Despite this misspecification, the similarity between probit and logit distributions preserves meaningful information in the parameter estimation, causing the weighting scheme to still favor more informative model specifications.
Furthermore, Figure \ref{fig:sum of weights} plots the sum of weights assigned to uninformative models ($\|\check{\boldsymbol{w}}\|_1$) under Case 2. As evident from the figure, the weight sum converges to zero as the sample size increases, which agrees with the theoretical property in Theorem \ref{theorem:weights convergence}.

In addition to the above, we conducted extensive simulations across additional scenarios to systematically evaluate the robustness of the proposed method. First, in Case 2 of this section, the sources contain models identical to the true parameters of the target. Given that real-world parameters exhibit similarity rather than identity, we formalize this phenomenon in Supplement Material. As demonstrated by the simulation results, the proposed model continues to perform well. Subsequently, we further investigated the impact of additional potential factors, including the target sample size $n_0$, the proportion of known labels $p_L$ and the number of variables $p$ – where high-dimensional regimes are of particular interest. The results show that the proposed method can still effectively leverage information from other models to assist in target prediction. The specific results are presented in Supplement Material.

\subsection{Application}
\label{sec: application}
To validate the effectiveness of the proposed method using real-world data, we employ a peer-to-peer (P2P) lending data set obtained from a famous chinese online lending platform  established in 2010. The data set encompasses three critical machine learning scenarios: (1) explicit binary classification labels, where positive cases are defined as observations with default occurrences, and negative cases consist of samples with successful loans and timely repayments; (2) PU learning setting where only positive cases are recorded; and (3) abundant semi-supervised learning opportunities through numerous unlabeled samples that could potentially belong to either category. This multifaceted data structure accurately reflects the core challenges in real-world financial risk control: the inherent rarity of default cases renders conventional supervised learning approaches ineffective, while the substantial volume of unlabeled data contains latent risk patterns that necessitate integrated PU learning and semi-supervised methodologies for comprehensive pattern discovery.

Our study encompasses data sets from ten provinces, ranked in descending order of sample size. The smallest data set, from Heilongjiang, has 507 samples with the proportion of positive labeled samples $p_L=0.21$, and the rest are unlabeled samples. This is designated as the target PU domain, and the remaining nine provinces are treated as source domains, with sample sizes for each province listed as follows:
\begin{itemize}
\tightlist
    \item Guangdong (7399 samples), Jiangsu (5693 samples), and Henan (3552 samples) have sufficient fully labeled binary classification data, enabling the training of relatively accurate classifiers using complete labels; these are designated as binary classification data sources.
    \item Hebei (2704 samples), Beijing (2266 samples), and Yunnan (1300 samples) have relatively limited labeled samples, necessitating the inclusion of unlabeled data for auxiliary estimation; these are treated as semi-supervised data sources.
    \item Sichuan (945 samples, $p_L=0.47$), Anhui (879 samples, $p_L=0.43$) and Zhejiang (600 samples, $p_L=0.17$) contain adequate default samples but lacked verified negative samples, and are therefore designated as PU data sources.
\end{itemize} 
The analysis incorporates \( p = 13 \) variables, with detailed descriptions provided in Supplement Material. Notably, the continuous variables are scaled to ensure transferability.  

\begin{figure}[H]
    \centering
    \includegraphics[width=0.8\linewidth]{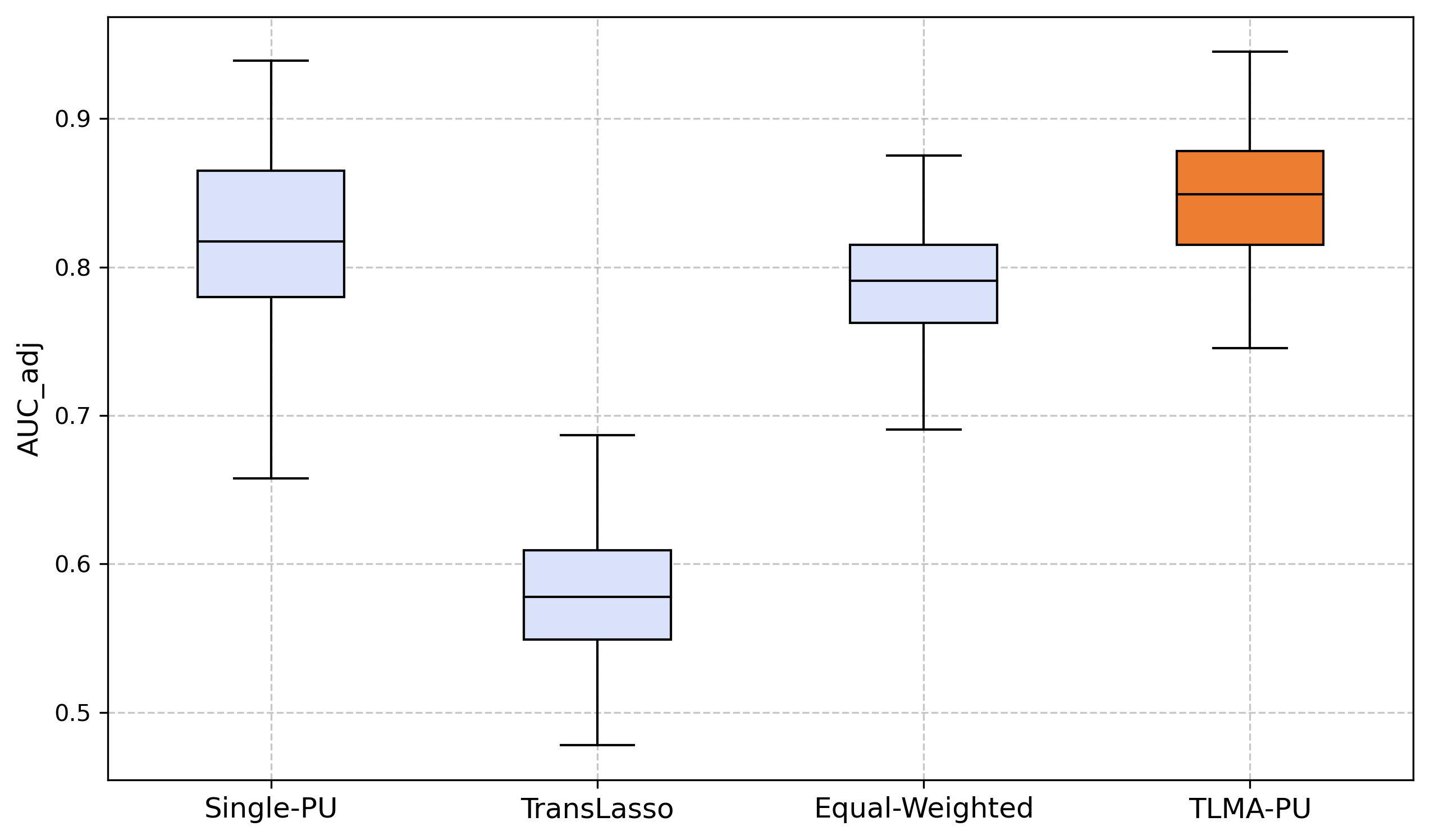}
    \caption{The AUC\_adj of compared methods.}
    \label{fig:app_aucadj}
\end{figure}

The data set is randomly partitioned into training and testing sets over 100 repetitions, with 80\% of the data used for training and the remaining 20\% reserved for testing. The AUC\_adj of the compared methods is calculated, except for the Oracle method, which is not applicable in this section. 
The results are presented in Figure \ref{fig:app_aucadj}.
From the results, it is evident that the proposed method performs well, achieving a relatively high overall AUC\_adj, indicating its effectiveness in handling the given task. 

\section{Conclusion}
\label{sec: conclusion}
This study introduces model averaging into PU learning and achieves privacy-preserving transfer learning for PU data. 
First, the source domain data configuration is flexible, accommodating fully labeled binary data, semi-supervised data, or PU data. For each data type, corresponding likelihood functions are constructed to estimate parameter.
Subsequently, we employ model averaging for transfer learning, which enables parameter transmission between data sources without directly sharing raw data or gradients that may compromise privacy.
Theoretically, we establish the optimality of weight selection under model misspecification for both in-sample and out-of-sample scenarios. When the target model is correctly specified, we prove the convergence of the weights. Moreover, these results are extended to high-dimensional settings, ensuring the robustness and applicability of the framework in more complex scenarios. Extensive numerical simulations evaluate the impacts of key factors including target or source sample sizes, predictor dimensionality, and labeled proportion. The proposed method demonstrates robust performance across all comparative analyses. Empirical applications further validate its effectiveness in default prediction tasks.
Notably, the target domain is not limited to PU data, it can also extended to semi-supervised data. Under certain assumptions, the modified logistic regression likelihood function can be provided for the corresponding data types, and the appropriate weight selection criteria can then be established.

Several aspects of this study warrant further investigation. First, in practical applications, data distributions may exhibit temporal dynamics. Developing adaptive transfer learning algorithms capable of adjusting dynamically to such distributional shifts presents an intriguing research direction. Furthermore, differential privacy techniques offer a promising approach to ensure privacy preservation while addressing data heterogeneity issues. Integrating these techniques with federated learning frameworks, particularly through the strategic injection of calibrated noise during model training, could enhance model generalizability while rigorously maintaining data privacy guarantees.

% \begin{itemize}
% \tightlist
% \item
%   Note that figures and tables (such as Figure~\ref{fig-first} and
%   Table~\ref{tbl-one}) should appear in the paper, not at the end or in
%   separate files.
% \item
%   In document preamble, you may set the key \texttt{anon} to ``0'' to hide the authors and acknowledgements,
%   producing the required anonymized version. Set the key \texttt{anon} to ``1'' to produce the manuscript with author details and acknowledgments. 
% \item
%   Remember that in the anonymized version, you should not identify
%   authors indirectly in the text. That is, don't say ``In Smith et.
%   al.~(2009) we showed that \ldots{}''. Instead, say ``Smith et.
%   al.~(2009) showed that \ldots{}''.
% \item
%   These points are only intended to remind you of some requirements.
%   Please refer to the instructions for authors at
%   \url{http://amstat.tandfonline.com/action/authorSubmission?journalCode=uasa20&page=instructions\#.VFkk7fnF_0c}
% \item
%   For more about ASA~style, please see
%   \url{https://files.taylorandfrancis.com/asa-style-guide.png}.
% \item
%   If you have supplementary material (e.g., software, data, technical
%   proofs), identify them in the section below. In early stages of the
%   submission process, you may be unsure what to include as supplementary
%   material. Don't worry---this is something that can be worked out at
%   later stages.
% \end{itemize}

\section*{Disclosure statement}\label{disclosure-statement}

The authors have no conflicts of interest to declare.

\section*{Data Availability Statement}\label{data-availability-statement}

Due to commercial confidentiality agreements, the data supporting this study are not publicly available.

\bibliography{bibliography.bib}

% \newpage
% \phantomsection\label{supplementary-material}
% \bigskip

% \begin{center}

% {\large\bf SUPPLEMENTARY MATERIAL}

% \end{center}

% \begin{description}
% \item[Title:]
% Brief description. (file type)
% \item[R-package for MYNEW routine:]
% R-package MYNEW containing code to perform the diagnostic methods
% described in the article. The package also contains all datasets used as
% examples in the article. (GNU zipped tar file)
% \item[HIV data set:]
% Data set used in the illustration of MYNEW method in
% Section~\ref{sec-verify} (.txt file).
% \end{description}
\end{document}